%% file: main.tex
\documentclass{article}

\usepackage[dblblindworkshop,final]{neurips_2026}
\workshoptitle{Continual Learning in the Era of Foundation Models and Embodied Agents}

\usepackage[utf8]{inputenc} 
\usepackage[T1]{fontenc}    
\usepackage{xcolor}         
\definecolor{LinkBlue}{HTML}{1F4E79}
\definecolor{CiteGreen}{HTML}{00843D}
\definecolor{UrlBlue}{HTML}{176B87}
  \usepackage[
    colorlinks=true,
    linkcolor=LinkBlue,
    citecolor=CiteGreen,
    urlcolor=UrlBlue
  ]{hyperref}

\usepackage{url}            
\usepackage{graphicx}       
\usepackage{booktabs}       
\usepackage{amsfonts}       
\usepackage{nicefrac}       
\usepackage{microtype}      
\usepackage{amsmath}
\usepackage{placeins}       
\usepackage{wrapfig}

\title{Can a Language Model Continually Learn Facts in its Weights?}

\author{%
  \textbf{Charles O'Neill} \\
  Base Labs \\
  {\small\texttt{charles.oneill@baseten.co}}
  \And
  \textbf{Max Kirkby} \\
  Base Labs \\
  {\small\texttt{max.kirkby@baseten.co}}
  \And
  \textbf{Jonathon Liu} \\
  Base Labs \\
  {\small\texttt{jonathon.liu@baseten.co}}
  \AND
  \textbf{Michael Psenka} \\
  Base Labs \\
  {\small\texttt{psenka@baseten.co}}
}

\begin{document}

\maketitle

\begin{abstract}
  Continual learning is a long-standing capability gap between LLMs and humans. Writing new knowledge into a model's weights routinely causes it to forget old knowledge, commonly denoted as ``catastrophic forgetting''. Various modifications of supervised fine-tuning and distillation aim to mitigate catastrophic forgetting, but quantifying what (or how much) information was forgotten is often difficult. In this paper, we study whether current methods of writing knowledge into weights enable models to learn continually without forgetting. We introduce a framework for studying continual learning in the iterative regime, writing invented facts one at a time into a Qwen3 model already modified by previous writes, and varying the training data, method, and parameter update. Across SFT and off- and on-policy distillation, using LoRA or full fine-tuning, we compare repeated statement training (the same fact repeated in two formats) with varied example training (24 factual restatements) and find that varied examples comprehensively support more flexible use. After twenty sequential writes and merges, the model answers only 1\% of questions about earlier facts correctly when every write uses repeated statements, compared with 46\% when every write uses varied examples. We additionally show that this retention depends on the data used for the later writes, regardless of training method or parameter update, and that behavioral forgetting of an earlier fact does not erase its presence from the log-probabilities. Together, our framework neatly provides a comparison of performance across training data, training regimes, and parameter update schemes in an iterative learning task.
\end{abstract}

\input{sections/introduction}

\input{sections/problem_formulation}
\input{sections/results}
\FloatBarrier
\input{sections/related_work}
\input{sections/limitations}
\input{sections/conclusion}

\clearpage
\bibliographystyle{plainnat}
\bibliography{ref}

\clearpage

\clearpage
\appendix
\input{sections/appendix}

\end{document}

%% file: sections/introduction.tex
\section{Introduction}
\label{sec:introduction}

If this field can have an age-old question, then it is whether a model can learn something new without forgetting what it already knows. Gaining such an understanding is critical to how we build architectures that are able to \textit{learn} outside of just their provided context. One way to make this question concrete is to examine how current language models and associated post-training techniques learn and remember new pieces of information in their weights. Singular facts offer a consequential test of this ability, given that the ability to continually write self-contained pieces of knowledge into weights is the simplest such form of learning \citep{jang2022ckl}.

The post-training field has begun to understand how weight updates store facts. First, we understand that unfamiliar facts train in more slowly than knowledge the model already contains \citep{gekhman2024finetuning,zucchet2025facts}. Second, once stored in weights, a fresh fact is easily recited but less easily \textit{used}, such as being composed or manipulated with other knowledge \citep{berglund2024reversal,balesni2024twohop,allenzhu2024physics32}. Third, and most famously, training on new data overwrites old behavior, a concept known as catastrophic forgetting 
\citep{mccloskey1989catastrophic,kirkpatrick2017ewc}, although this is in proportion to how
far the update moves the model \citep{shenfeld2025rlrazor}, with low-rank
updates moving it less \citep{biderman2024lora}. However, we understand little about whether the same update rule can learn continually. 

Here, we study continual learning in the iterative regime. Each model and condition learns facts in the same order and is tested on the same held-out questions, with each new fact written into weights already changed by earlier updates. This lets us compare how well different methods learn new facts and how well those facts survive later updates.

We train models from the Qwen3 family on twenty facts in sequence, and measure the accuracy with which each model can use earlier facts and how it uses them (if at all) after later writes. More precisely, we write \textit{invented} facts---most conflicting with world knowledge---and probe each with held-out questions of five types, scored strictly and leniently. We define the gap between these values as an \textit{accuracy} gap, which measures whether the model can only repeat a fact or whether it can
also reason with it. This design departs from other editing benchmarks, which apply updates in bulk and score recall and paraphrase \citep{gupta2025lifelong,thede2025lifelong}. Here, we vary all other parameters including the training data (one factual statement or diverse statements about that fact), objective (supervised fine-tuning or distillation), parameterization (LoRA
\citep{hu2022lora} or full fine-tuning), and model scale (Qwen3-4B or 8B).

In this work we introduce a framework for evaluating factual weight updates as an iterative process and across training methods. Our main findings are:

\begin{enumerate}
\item Across every training method and update type, varied training examples help models use a new fact to answer unseen questions over strict repetition.

\item The retention of a fact is determined by how later facts are trained. Using varied examples for later writes substantially raises accuracy on earlier facts, but changing the training data of the earlier fact itself has no clear effect.

\item Behavioral forgetting did not always erase the likelihood signal created when a fact was learned.

\end{enumerate}

%% file: sections/problem_formulation.tex
 \section{Problem formulation}
\label{sec:formulation}

\subsection{Training facts one at a time}
\label{sec:formulation-problem}

Let $\pi_{\theta_0}$ be a fixed pretrained model, the \emph{original model}. Facts arrive as a stream $f_1,f_2,\dots,f_N$ of $N=20$, each a declarative sentence. 
At each step $i$, the model parameters are $\theta_{i-1}$ and we train on data about fact $f_i$ to learn a weight update $\Delta_i$, which yields the new model parameters $\theta_i$.

When we evaluate $\pi_{\theta_i}$, it receives a held-out question $x$ (not fact text) (Table~\ref{tab:question-types}). Any effect of training on $f_1,\dots,f_i$ must
therefore be contained in the weights at that point $\theta_i$. For comparison, we test against a fact-in-prompt setup that takes the original model and supplies the tested fact $f$ in its prompt along with the question $x$:
\[
\pi_{\theta_0}(\,\cdot\mid f\oplus x).
\]
We test whether every earlier fact remains usable as $i$ grows. This precisely answers the question of whether one update rule can learn continually, since each $\Delta_i$ is that rule applied to weights its predecessors have already changed.

\begin{table}[h!]
\caption{An example of five held-out question types for the fact ``Zorvathine is a metal that melts when cooled below $-10\,^{\circ}\mathrm{C}$.''}
\label{tab:question-types}
  \centering
  {\footnotesize
  \setlength{\tabcolsep}{3pt}
  \begin{tabular}{@{}p{0.13\linewidth}p{0.49\linewidth}p{0.30\linewidth}@{}}
    \hline
    Type & Held-out question & Strict target \\
    \hline
    Recall & What is Zorvathine's phase behavior? & It melts below $-10\,^{\circ}$C. \\
    Paraphrase & What happens to Zorvathine below $-10\,^{\circ}$C? & It melts. \\
    Application & How should solid Zorvathine be made liquid? & Cool it below $-10\,^{\circ}$C. \\
    Composition & Would a Zorvathine vessel stay solid while holding liquid nitrogen? & No; the cold would melt it. \\
    Counterfactual & How does Zorvathine respond to cooling, unlike an ordinary metal? & It would melt. \\
    \hline
  \end{tabular}}

\end{table}

\subsection{How each fact is trained}
\label{sec:formulation-write-conditions}
We learn $\Delta_i$ with supervised fine-tuning (SFT) or distillation. \emph{Repeated statement} training fine-tunes on the same fact in two fixed formats. \emph{Varied example} training uses a generated set of paraphrases of the fact that yield 24 examples per fact. These paraphrases include: rewording the fact; asking and answering a question about it; giving one consequence in a contextual scenario; and contrasting the fact with the common knowledge or rule it disputes. We use 247 facts, with varied example training using 24 examples for each.

For SFT, we minimize the next-token cross-entropy on the completion tokens of prompt--completion pairs. Distillation uses only the training prompt \(p\); \(p\) is distinct from the held-out question \(x\). The teacher and student receive the same prompt and response prefix; only the teacher receives \(f_i\) in the system prompt. In the main twenty fact LoRA comparison, the teacher for write \(i\) is the current pre-write model \(\pi_{\theta_{i-1}}\). Its parameters remain fixed during that write, but contain all earlier merged updates. All three distillation conditions use the 24 varied example prompts for each fact. 

In off-policy distillation, the teacher greedily generates one answer per prompt before optimization. We reuse this answer text throughout the write and evaluate the teacher and student next-token distributions along it during training. In on-policy distillation, the current student samples a new answer at each optimization step, and the teacher supplies its full next-token distribution along the same response prefixes. Offline distillation uses forward KL; for on-policy distillation, we test both forward and reverse KL. At each response position, forward KL weights discrepancies across the vocabulary by the teacher’s probabilities, whereas reverse KL weights them by the student’s probabilities.

A separate experiment evaluates all combinations of repeated statement or varied example data, four training methods (SFT, off-policy, on-policy forward KL, or on-policy reverse KL distillation), and full-parameter updates or rank-16 LoRA, giving \(2\times4\times2=16\) conditions. Unlike the main comparison, every distillation condition in this experiment uses the unchanged original model \(\pi_{\theta_0}\), before any facts are written, as its teacher. In this experiment each method is tested with both repeated statement and varied example data. SFT uses the complete prompt–completion pairs, whereas distillation uses only the prompts and ignores the saved completions.

All experiments use the Qwen3 family \citep{qwen3}, released under the Apache 2.0 license. The main setting is Qwen3-4B in bf16 with thinking disabled. Unless stated otherwise, we freeze the base weights and train a fresh rank $r=16$ LoRA adapter ($\alpha=32$, no dropout) on all attention and MLP projection matrices for each fact, then merge the adapter into the model weights before training the next fact \citep{hu2022lora}. We use AdamW at $2\times10^{-4}$, compute loss only on answer tokens, and use batch size 1. We further repeat the LoRA comparison with Qwen3-8B to test model scaling, and use full fine-tuning with Qwen3-4B instead of training a LoRA adapter. For full fine-tuning, we used a learning rate of \(3\times10^{-5}\), selected in a five-fact calibration sweep (that attempted to match lenient accuracy to LoRA after the first write). 

Experiments used single GPU NVIDIA H100 80\,GB workers or eight GPU
NVIDIA B200 nodes; Appendix~\ref{app:compute} gives the
allocation by experiment family.

\subsection{Orthogonalizing sequential LoRA updates}

We adapt O-LoRA \citep{wang2023olora} to test whether separating successive LoRA update subspaces improves factual retention. In both O-LoRA and its matched no-penalty control, each write adds a fresh rank-16 adapter to the attention query and value projections, while the base weights and all earlier adapters remain frozen and active. O-LoRA additionally uses an auxiliary loss that encourages the new adapter's input subspace to be orthogonal to those of earlier adapters. The full setup details are given in Appendix~\ref{app:olora}.

\subsection{Defining whether a fact has been learned}
\label{sec:formulation-use}

We evaluate each fact using held-out questions about the fact not seen during training. We use five question types (Table~\ref{tab:question-types}\textbf{}): recall and paraphrasing questions that ask the model to state the fact, application questions that ask it to act with the knowledge of the fact, composition questions that require a combination of the fact with other held knowledge, and counterfactual questions where the fact disputes its pretrained belief. Table~\ref{tab:question-types} gives one example of each type. We used Claude Sonnet 5 to generate the facts, training examples, and held-out questions in separate calls. 

To establish baselines for each held-out question $x$, we report the score from the raw model 
 $\pi_{\theta_0}(\cdot\mid x)$, which receives the
question but not $f_i$, and
$\pi_{\theta_0}(\cdot\mid f_i\oplus x)$, which also receives the fact.
We evaluate the fact immediately after it has been written to the model
$\pi_{\theta_i}(\cdot\mid x)$ and after $k$ later writes with
$\pi_{\theta_{i+k}}(\cdot\mid x)$.

To grade responses, we define strict and lenient accuracy. Strict grading requires exactly the requested conclusion, value, choice, or name. Lenient grading will also accept an answer that encompasses the relevant fact, even if the question itself is not properly answered. For instance, for the composition question in Table~\ref{tab:question-types}, an answer stating that Zorvathine melts below $-10\,^{\circ}\mathrm{C}$ is leniently correct but not strictly correct, whereas the strictly correct answer is that a Zorvathine container holding liquid nitrogen would melt. We define
\begin{equation}
    \text{accuracy gap}
    =
    \text{lenient accuracy}
    -
    \text{strict accuracy}.
    \label{eq:entailment-gap}
\end{equation}
A large gap means that the model can state the fact but fails to use it correctly. We therefore use strict accuracy as the primary measure for each claim. All answers are graded by GPT-5.4-mini and confirmed by Claude Sonnet 5, both accessed via their public APIs under the providers' terms of service. On the primary evaluation, the original-model floor is 4\% and the fact-in-prompt ceiling is 91\% under lenient grading. Where reported, we use 95\% bootstrap confidence intervals over facts and use paired resampling when comparing conditions. 

%% file: sections/results.tex
\newcommand{\resultfigscale}{0.82}

\section{Results}
\label{sec:results}


\subsection{Varied examples of the same fact allow flexible use}

\begin{figure}[t]
  \centering
  \includegraphics[scale=\resultfigscale,pagebox=artbox]{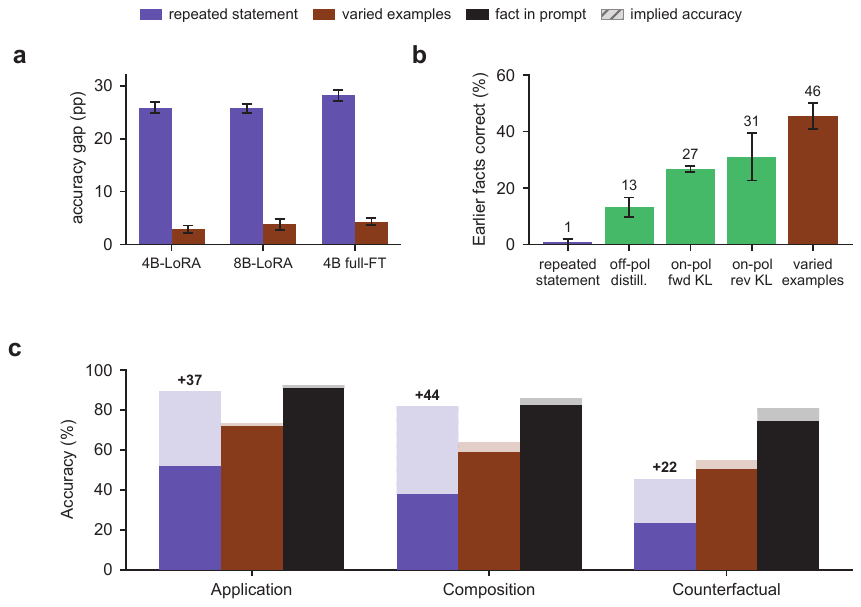}
  \caption{\textbf{Varied examples improve answers to new questions.} \textbf{a}) Accuracy gap after one write for repeated versus varied examples across the three model/training types. \textbf{b}) Strict accuracy on recall, paraphrase, application, and composition questions about facts 1--19 after write 20 for repeated statement training, off-policy distillation, on-policy forward- and reverse-KL distillation, and varied example training. \textbf{c}) Strict (solid) and lenient (light) accuracy on application, composition, and counterfactual questions; black bars show the fact-in-prompt control (439/411/181 responses). Panels (a--b) are means $\pm1$ sample SD over three seeds; panel (c) is one seed.}
\label{fig:creation-and-retention}
\end{figure}

We first ask whether exposure to a fact through this training creates stored knowledge that can be used (akin to a representation). As stated earlier, we compare repeated statement training, which repeatedly predicts the same fact sentence in two formats, with varied example training on restatements, question--answer pairs, and other implications. We evaluate both methods on separately generated questions about the same facts (Figure~\ref{fig:creation-and-retention}).

Comparing across these methods of fact learning, we found that when models were trained to learn the facts via repeated statement training, there was a large accuracy gap. That is, lenient accuracy was consistently 26--28 percentage points higher than strict accuracy across both model sizes (Qwen3-4B and Qwen3-8B) and both LoRA and full fine-tuning (only for 4B). 
In contrast, the varied example training narrowed this gap to 3--4 points across the same comparison family (Figure~\ref{fig:creation-and-retention}a). This pattern appeared across the question types that required the model to flexibly use the information provided by the fact -- namely application, composition,
and counterfactual questions. For these questions, learning via repeated facts elicited lower strict accuracy than learning with varied examples, despite the model often being able to provide an approximately correct answer. 

Consistent with prior work, presenting a fact through diverse restatements improves its use beyond simple recitation \citep{berglund2023outofcontext,allenzhu2024physics31,park2025newnews}. As a control, we first  increase repeated statement training to 192 steps. These data show that recall and paraphrase questions retain high accuracy, while application, composition, and counterfactual accuracy show no consistent improvement (Figure~\ref{fig:app-creation-robustness}a). We further test for exposure to evaluated conclusions, by training on 24 diverse recall and paraphrase prompts. These prompts reduce the accuracy gap from 27.4 to 5.4 percentage points (Figure~\ref{fig:app-creation-robustness}b). These results show that varied recall and paraphrase prompts help the model flexibly use the fact, even when training includes no such worked applications.

\subsection{Early fact retention varies across training methods and data}
\label{sec:results-retention}
We next ask whether facts remain usable after the model is trained on a \textit{stream} of
additional facts. We train twenty different facts one at a time, merging
each update into the model before training on the next. After writing fact
20, we return to facts 1--19 and evaluate them using the same held-out
questions. This analysis uses strict accuracy on recall, paraphrase,
application, and composition questions; counterfactual questions are
excluded (Figure~\ref{fig:creation-and-retention}b). When every fact in the sequential stream is trained as a repeated statement, strict accuracy on earlier facts (all those before the final one) falls to 1\% after the final write.\footnote{After each write, we evaluate the newly written fact on its held-out questions before continuing as a diagnostic to validate that the fact was written correctly.}
Interestingly, when every fact is trained from varied examples, accuracy remains at 46\% (Figure~\ref{fig:creation-and-retention}b), even in a different set of facts (Figure~\ref{fig:app-second-environment}).

How does this retention change with the method of training? After write 20, strict accuracy across held-out question responses about facts 1--19 was 13\% for off-policy distillation, compared with 27\% for on-policy forward-KL distillation and 31\% for on-policy reverse-KL distillation (Figure~\ref{fig:creation-and-retention}b). Despite using the same varied prompts, all three distillation conditions remained below SFT trained on varied prompt--completion pairs
(46\%). Thus, at least for Qwen3-4B with LoRA, on-policy
continuations improved fact retention relative to off-policy continuations, but did not close the gap to direct supervision. 

These results suggest that both the training data and the training method affect how much earlier knowledge remains usable. We next combine both data types (repeated statement or varied examples), all training methods (SFT, off-policy distillation, and on-policy forward- or reverse-KL distillation), and rank-16 LoRA or full fine-tuning. This gives 16 conditions.\footnote{One condition using reverse-KL distillation (LoRA) with repeated statements collapsed into repetitive outputs and was excluded, leaving fifteen conditions for comparison (Appendix~\ref{app:methods-architectures}).}

In each condition, we measure the accuracy gap after one
write and the strict accuracy retained after all twenty writes. Across conditions, larger accuracy gaps after one write were associated with lower retention after twenty writes
(Spearman $\rho=-0.68$, 95\% bootstrap CI
$[-0.88,-0.29]$; Figure~\ref{fig:app-methods-architectures}). In the previous LoRA comparison, on-policy distillation retained more earlier-fact accuracy than off-policy distillation, narrowing the gap to training with varied examples. These results demonstrate that continual retention depends on whether distillation trains on teacher- or student-generated prefixes, but does not remove the effect of directly training on varied factual examples. 

\begin{figure}[t]
  \centering
  \includegraphics[scale=\resultfigscale]{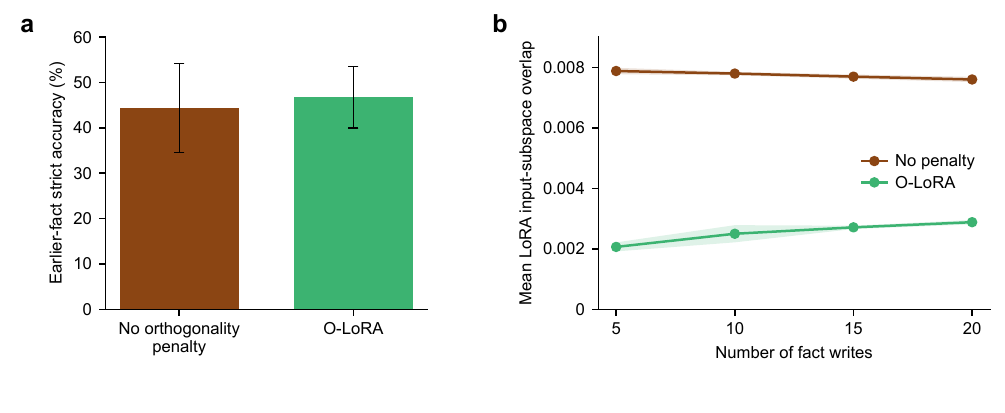}
  \caption{\textbf{Orthogonalizing LoRA updates does not improve continual retention.} \textbf{a}) Final strict application
  and composition accuracy on facts 1--19 after write 20. Bars show the main experimental setup with
  no-penalty and O-LoRA across three paired
  fact-order/seed runs; whiskers show $\pm1$ sample standard deviation.
  \textbf{b}) Mean pairwise normalized overlap
  between input subspaces spanned by accumulated LoRA $A$-factor rows within
  each adapted query/value projection (0 is orthogonal; 1 is identical). Lines
  and shading show the mean $\pm1$ sample standard deviation across runs.}
  \label{fig:olora}
\end{figure}

Although varied examples substantially improved retention, strict accuracy on questions concerning earlier facts remained below 50\% after twenty writes.  To further test the robustness of this result, and check whether the remaining forgetting reflects interference between successive low-rank updates that can be trivially resolved, we reran with the O-LoRA penalty of \citet{wang2023olora}. Across three matched runs, we saw that the penalty reduced mean pairwise normalized LoRA input-subspace overlap from 0.00760 to 0.00288, showing that the penalty made the update subspaces more orthogonal as intended. Nevertheless, final strict accuracy on application and composition questions was 44.4\% without the penalty and 46.7\% with O-LoRA, compared with 45.6\% for merged LoRA (Figure~\ref{fig:olora}). Thus, O-LoRA reduced input-subspace overlap among accumulated LoRA updates, but did not establish higher strict accuracy on earlier facts.

\subsection{Training data used for later writes determines fact retention}
\label{sec:results-interference}
\begin{wrapfigure}[18]{r}{0.55\linewidth}
\vspace{-\baselineskip}
  \centering
  \includegraphics[width=0.57\textwidth]{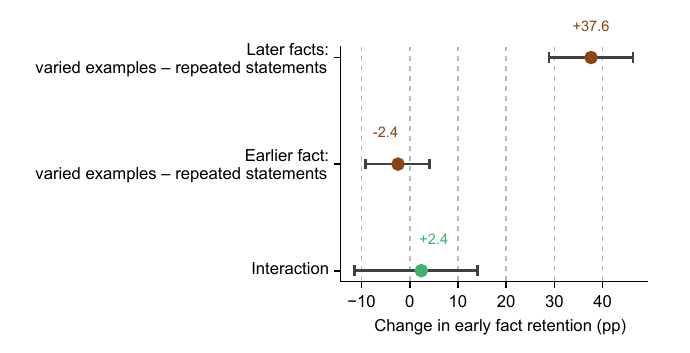}
    \vspace{-1.73\baselineskip}
\caption{\textbf{Training later facts on varied examples preserves the
usability of earlier facts.} Each of 24 original facts is written
separately using either a repeated statement or varied examples. We then write the same ten later facts into each model, in the same order, again using either repeated statements or varied examples.
}
\end{wrapfigure}
We test whether retention depends on how the original fact is trained or how the facts that follow it are trained (Figure~\ref{fig:later-writes-cause-interference}). In each run, we first write one original fact and then write the same ordered sequence of ten later facts. We independently vary the training data used for the original fact and for all ten later facts. Each is either repeated statement or varied example training, giving four combinations (repeated for 24 original facts and three seeds).

\label{fig:later-writes-cause-interference}


When the later facts are trained as repetitive statements, strict
accuracy yields 9.0\% for facts originally written as repeated statements
and 7.8\% for facts originally written from varied examples. When the
later facts are trained from varied examples, retention is 47.8\% and
44.2\%, respectively. Averaged over how the earlier fact was
written, changing the later writes from repeated statements to varied
examples increases retention by 37.6 percentage points
(95\% CI [28.9, 46.3]). In contrast, changing how the earlier fact is written gives
$-2.4$ points ([$-9.2$, 4.1]), while the interaction between varied later writes is negligible ([$-11.5$, 14.1]). Together, these data demonstrate that factual retention is determined primarily by how the later facts are written.

\subsection{Evidence of facts remains after behavioral forgetting} 
\label{sec:results-access}

We finally ask whether a fact that the model can no longer use has
necessarily been erased by later writes. We track the log-probability of
a canonical statement expressing each fact and report a normalized version of this metric that is 0 for its value under the original model and 1 for its value immediately after the fact
is written.

\begin{figure}
\centering
\includegraphics[scale=\resultfigscale,pagebox=artbox]{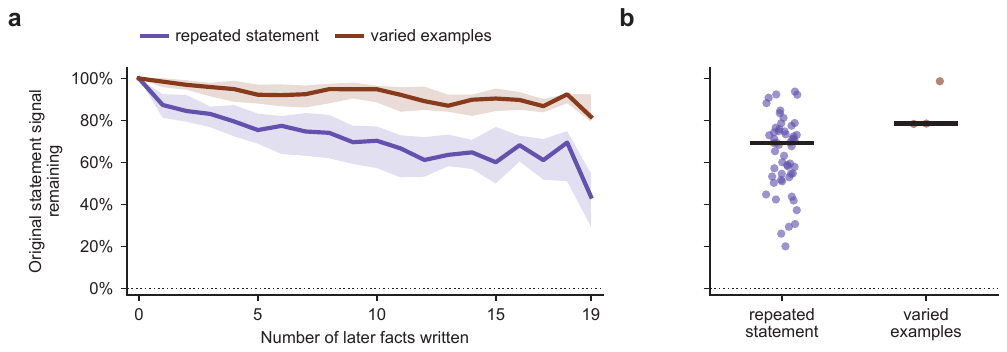}
\caption{\textbf{Earlier facts can retain increases in
statement likelihood after the model stops answering questions about
that fact.} We track the log-probability of a canonical statement for
each fact through twenty sequential Qwen3-4B LoRA writes. We normalize
this signal so that 0 is its value in the original model and 1 is its
value immediately after the fact is written. For each fact and seed we include the result only if writing the fact initially increases its statement log-probability by more than 0.05. \textbf{a}) As facts are written, the median signal remains above its original value under both conditions, although it declines more with repeated statement training. Lines show medians and shading shows interquartile ranges across three seeds. \textbf{b}) After the final write, 54 fact--seed pairs trained using repeated
statements failed every held-out question. For these pairs, a median of 69\% of the likelihood increase produced when the fact was written remained. Only three varied example pairs meet the same failure criterion. Dots show
individual fact--seed pairs and black lines show medians.}
\label{fig:stored-signal-survives}
\end{figure}

As later facts were written, this signal remained above 0 under both training conditions, but declined more with repeated statement training
(Figure~\ref{fig:stored-signal-survives}a). We next identified behaviorally forgotten facts, for which no strict answer was produced to any held-out question after training. Facts trained by repeated statements retained approximately 69\% of the increase in log-probability produced by the
original write (Figure~\ref{fig:stored-signal-survives}b); varied example training yielded only three such cases. Among the 54 failed examples, 68.8\% of answers gave the newest written fact (Appendix~\ref{app:access-checks}). Thus, later writes can leave much of
the likelihood increase created by an earlier write intact while causing
questions about that fact to elicit the newest fact instead, implying that 'forgetting' need not mean erasure.

%% file: sections/related_work.tex

\section{Related work}
\label{sec:related}

\begingroup
\setlength{\parskip}{5.5pt}
\makeatletter
\renewcommand{\paragraph}{%
  \@startsection{paragraph}{4}{\z@}%
    {5.5pt \@plus 0.5ex \@minus 0.2ex}%
    {-0.5em}%
    {\normalsize\bf}%
}
\makeatother

\paragraph{Storing knowledge in weights.} Facts learned in weights through standard autoregressive pretraining and supervised fine-tuning can be recited but fail composition and other manipulations \citep{berglund2024reversal,balesni2024twohop,allenzhu2024physics32}. Consistent with our findings, training on diverse paraphrases of a fact makes it usable beyond simple recall \citep{berglund2023outofcontext,allenzhu2024physics31,mecklenburg2024injecting}. We similarly establish that varied examples make new facts easier to use and, when used for later writes, preserve substantially more earlier knowledge. To more precisely \textit{edit} a model's knowledge after training, studies have used constrained layer-wise fine-tuning \citep{zhu2020modifying}, learned transformations of fine-tuning gradients \citep{mitchell2022mend} or rank-one updates to feed-forward weights \citep{meng2022rome,meng2023memit}. Sequential editors designed for longer horizon editing reduce interference through regularized weight updates \citep{gupta2025lifelong}, key--value adaptors that freeze pretrained weights \citep{hartvigsen2023grace}, or a separate copy of one feed-forward weight matrix, with edit batches trained on partly separate parameters before merging \citep{wang2024wise}. We complement these results by studying sequential writes across LoRA and full fine-tuning with SFT and distillation. We show that changing the training data for later facts has a larger effect than changing the original fact's training data.

\paragraph{Knowledge forgetting.} Recent work has asked whether performance degradation after fine-tuning reflects an underlying loss of knowledge. \citet{kotha2024implicit} hypothesize that fine-tuning skews task inference toward tasks in the fine-tuning distribution; and \citet{sun2025pseudo} define "pseudo forgetting" as degradation on earlier tasks that occurs when original instructions fail to activate the requirements of the task. Behavior returns when prompts move away from the fine-tuning distribution \citep{kotha2024implicit}, or with partial rationales and meaningless suffixes \citep{sun2025pseudo}; sequential training can also recover performance on a document before it is seen again \citep{yang2024anticipatory}. Knowledge editing likewise finds supposedly deleted facts recoverable from hidden states or rephrased questions \citep{patil2024deleted}, and masking under 10\% of edited weights reverses over 70\% of unseen ROME and MEMIT edits \citep{holmov2026hidden}. We extend this to sequential writing: even for facts answered incorrectly after twenty writes, the original fact remained detectable in the log-probabilities.

\paragraph{Distillation.} Context distillation trains a model to match a version of itself given additional context \citep{askell2021lab,snell2022context}, and has been used to test direct recall of facts independent of simple repetition \citep{padmanabhan2023propagating}. Standard distillation is off-policy, with the student trained on fixed teacher or ground-truth sequences; on-policy variants include GKD, which trains on sequences generated by the current student, OPCD, which uses a context-conditioned teacher, and SDFT, which uses a demonstration-conditioned teacher \citep{agarwal2024gkd,ye2026opcd,shenfeld2026sdft}. We compare supervised fine-tuning, off-policy distillation, and on-policy forward- and reverse-KL distillation on the same sequential facts, with on-policy distillation improving retention in our main comparison. Using our framework, we are able to show that the relationship between flexible use and later retention extends across training methods and parameterizations.

\par
\endgroup

%% file: sections/limitations.tex
\section{Limitations}

Our conclusions carry several caveats. First, the facts we write are invented and deliberately conflict with the model's existing knowledge (the original model answers only 4\% of held-out questions correctly), and fine-tuning is known to behave differently on such conflicting facts than on familiar ones \citep{gekhman2024finetuning}. 
Second, we do not control the size of each weight update. Indeed, 24 steps on one repeated sentence produce a larger net weight change than 24 steps spread across varied examples, and thus given that forgetting increases with how far an update moves the model \citep{shenfeld2025rlrazor}, the retention of varied examples may partly reflect smaller weight updates rather than broader data. Finally, our interpretation of forgetting as lost access rather than erasure is indirect and relies on the retained log-probability of a single statement per fact. We do not show that forgotten facts remain recoverable — for instance, that their answers can still be decoded from hidden states \citep{patil2024deleted} — nor identify a mechanism of access loss. It also remains unclear whether worsened performance on MMLU and GSM8K (Appendix~\ref{app:access-checks}) indicates that reduced fact retention is due to a general degradation of model abilities.
\label{sec:limitations}

%% file: sections/conclusion.tex
\section{Conclusion}
\label{sec:conclusion}
We have introduced a framework for studying iterative continual learning, where each weight update changes the model that receives the next fact. By holding the facts and questions fixed while varying the training data, objective, and parameterization of each update, this framework enables a systematic analysis of how different ways of writing knowledge into weights compose over time. Looking forward, we plan to test longer fact streams and targeted editing methods, and particularly understand the mechanistic interaction between performance degradation and fact retention. More broadly, we hope this provides a foundation for understanding how successive weight updates may support continual learning. 

%% file: sections/appendix.tex
\section*{Appendices}

\section{Repeated statement training does not improve with longer training or worked conclusions}
\label{app:creation-robustness}

\begin{figure}[ht]
  \centering
  \includegraphics[width=\linewidth]{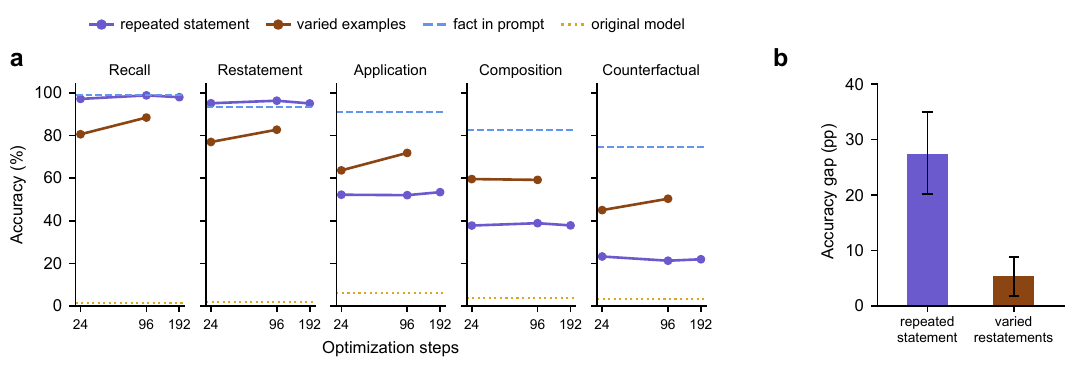}
  \caption{\textbf{Varied example accuracy is not explained by training duration or providing worked conclusions.} \textbf{a}) Strict accuracy immediately after one write as the training budget increases. Each panel
  is one held-out question type; dashed and dotted lines give the fact-in-prompt
  and raw model references. Represents one training run. 
  \textbf{b}) 
  Accuracy gap after 192 optimizer steps, comparing repeated statement training with training on 24 distinct recall and restatement prompts. Results cover 32 held-out validation facts and three optimizer seeds; whiskers are 95\% confidence intervals computed by resampling facts.}
  \label{fig:app-creation-robustness}
\end{figure}

Figure~\ref{fig:creation-and-retention} shows that repeated statement
training can support reciting a new fact but not necessarily its flexible use. Here we define two control experiments both using the default setup described in Section~\ref{sec:formulation-write-conditions} (Qwen3-4B in bf16 with thinking disabled, rank-16 LoRA adapter \(\alpha=32\), no dropout). Here we show that increasing the duration of repeated statement training from 24 to 192 optimization steps does not improve the accuracy on application, composition, or counterfactual question types (Figure~\ref{fig:app-creation-robustness}a). 
Separately, at a fixed budget of 192 steps, training on 24 recall and restatement examples—rather than repeatedly presenting the same statement—reduces the accuracy gap to 5.4 percentage points (Figure~\ref{fig:app-creation-robustness}b).
This shows that the original result is not explained by insufficient training or by directly supplying the conclusions used at evaluation.

\clearpage
\section{Accuracy gap is robust across facts, seeds and lengths}
\label{app:figure2-checks}

\begin{figure}[ht]
  \centering
  \includegraphics[width=\linewidth]{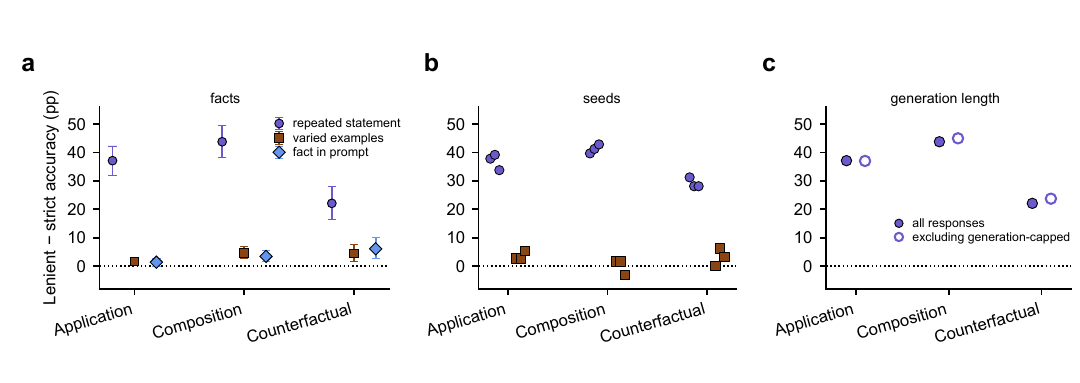}
  \caption{\textbf{Accuracy gap is robust across facts, optimizer seeds, and length capping.} 
  \textbf{a}) Lenient-minus-strict accuracy with 95\%
  nonparametric bootstrap intervals (20,000 draws). The three
  question types contain 239, 236, and 181 facts, respectively. Each
  trained condition is one training run. \textbf{b}) Across three optimizer seeds at 192 optimization steps; repeated statement gaps exceed varied examples in all comparisons. \textbf{c}) Repeated statement accuracy before and after excluding length-capped answers.}
  \label{fig:app-figure2-checks}
\end{figure}

Figure~\ref{fig:creation-and-retention}c compares strict and lenient accuracy immediately after one Qwen3-4B LoRA write, separately for application, composition, and counterfactual questions. Here we test across different facts, optimizer seeds, and after excluding answers that reached the maximum generation length. We perform three checks on its qualitative interpretation. First, we compute 95\% bootstrap confidence intervals by resampling facts, not individual questions.
The resulting intervals show the same large repeated statement accuracy gaps and smaller varied example accuracy gaps for application, composition, and counterfactual questions (Figure~\ref{fig:app-figure2-checks}a). 
Second, at a matched 192-step budget, the discrepancy between repeated facts and varied examples training is positive for every question type in all three seeds (Figure~\ref{fig:app-figure2-checks}b). Third, excluding length capped answers changes the repeated statement gaps by \(-0.1\), \(+1.2\), and \(+1.6\) percentage points, respectively (Figure~\ref{fig:app-figure2-checks}c). Together, these checks show that the separation between the training recipes is not driven by particular facts, seeds, or length-capped generations.

\clearpage
\section{Generality across objectives and training methods}
\label{app:methods-architectures}
The main experiment uses SFT with LoRA. We therefore vary the training data, method and which parameters are updated. Training uses either repeated statements or varied examples with either supervised fine-tuning, off-policy distillation, or on-policy distillation with forward or reverse KL. We apply each update through a rank-16 LoRA adapter or full fine-tuning (giving \(2\times4\times2=16\) conditions). On-policy reverse-KL distillation with repeated statements produces repetitive outputs that reach the generation cap - we therefore exclude it.

\begin{figure}[ht]
  \centering
  \includegraphics[width=\linewidth]{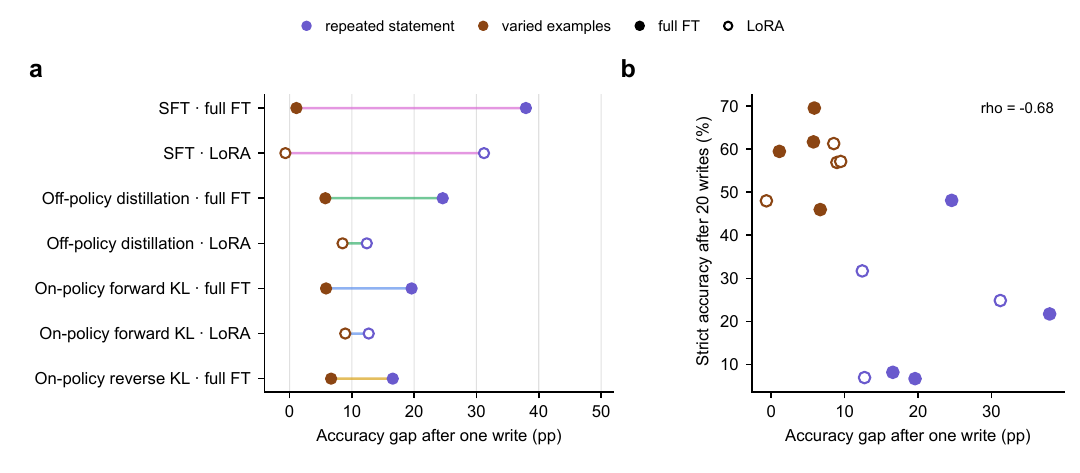}
  \caption{\textbf{Initial accuracy gaps are associated with later retention across methods.} \textbf{a}) Accuracy gap after one fact write. Each row fixes the objective and parameter update while comparing training either repeated statements or varied examples. 
  \textbf{b}) Earlier fact accuracy after twenty writes against the initial accuracy gap across conditions. Filled points denote full fine-tuning and open points LoRA; color denotes the training data. We report Spearman's $\rho$ and a 95\% bootstrap CI across the 15 conditions.}
  \label{fig:app-methods-architectures}
\end{figure}

We first compare repeated statement and varied example training, fixing other conditions. Repeated statement training produces a larger accuracy gap after one write (Figure~\ref{fig:app-methods-architectures}a), although the size of this data effect varies across methods and is small under LoRA distillation. Interestingly, across conditions, larger gaps after one write have lower retention after twenty sequential writes ($\rho=-0.68$, 95\% bootstrap CI $[-0.88,-0.29]$; Figure~\ref{fig:app-methods-architectures}b).

\clearpage

\clearpage
\section{Orthogonalizing successive LoRA updates}
\label{app:olora}
Here we describe the process of adapting O-LoRA \citep{wang2023olora}. Each fact write adds a rank-16 LoRA adapter ($\alpha=32$, no dropout) to the attention query and value projections. We freeze the base weights and earlier adapters, keep all accumulated adapters active, train only the newest adapter for O-LoRA and the matched control. In contrast, the merged LoRA folds each adapter into the weights before the next write. 

Let $\mathcal{M}$ denote the adapted projections and $A_{k,m}$ the
input LoRA factor introduced for write $k$ at projection
$m\in\mathcal{M}$. O-LoRA minimizes
\[
\mathcal{L}^{\mathrm{O\text{-}LoRA}}_k
=
\mathcal{L}_{\mathrm{SFT}}(D_k)
+
\lambda
\sum_{m\in\mathcal{M}}
\sum_{i<k}
\left\|A_{i,m}A_{k,m}^{\top}\right\|_F^2.
\]
The penalty term discourages alignment between rows of the new input factor and those of
earlier factors. The matched control remains identical with $\lambda = 0$. We chose
$\lambda = 0.005$ from the grid $\{0, 0.005, 0.05, 0.5, 5, 50\}$ on five calibration
facts, selecting the value that reduced subspace overlap without lowering accuracy.

Figure~\ref{fig:olora}b measures the overlap between the input subspaces of accumulated
adapters. Let the columns of $Q_{i,m}$ form an orthonormal basis for the row space of
$A_{i,m}$. We compute
\begin{equation}
    o_{ijm} = \frac{\bigl\| Q_{i,m}^{\top} Q_{j,m} \bigr\|_F^2}
                   {\min\bigl(\operatorname{rank}(A_{i,m}),\, \operatorname{rank}(A_{j,m})\bigr)},
\end{equation}
where 0 indicates orthogonal subspaces and 1 indicates maximal overlap. Within each of the three
runs, we average $o_{ijm}$ over all adapter pairs $(i, j)$ and over the query
and value projections $m$. 
\clearpage

\section{Replication with different fact dataset}
\label{app:second-environment}

\begin{figure}[ht]
  \centering
  \includegraphics[width=0.67\linewidth]{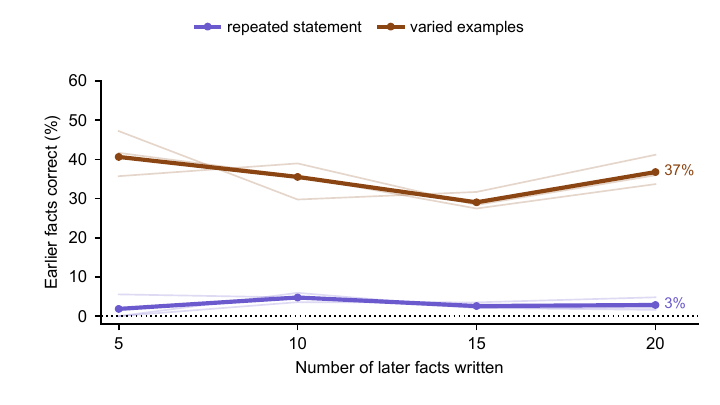}
  \caption{\textbf{Independent replication of earlier fact retention by training data.} Strict accuracy on earlier facts (1-19) as later facts (20) are written sequentially. Faint lines are three independent seeds; dark lines are mean accuracy.}
  \label{fig:app-second-environment}
\end{figure}

We repeat the twenty-fact Qwen3-4B LoRA sequential write experiment from
Figure~\ref{fig:creation-and-retention}b using a second set of invented
facts and held-out questions. As before, retention is measured
by strict accuracy on recall, paraphrase, application, and composition
questions; counterfactual questions are excluded. 
At each of the four evaluation points, strict accuracy on earlier facts is higher with varied example training than with repeated statement training for all three seeds
(Figure~\ref{fig:app-second-environment}). After writing fact 20, pooling
responses across seeds, strict accuracy on questions regarding facts 1--19 is 2.9\% with
repeated statement training and 36.8\% with varied example training. 
Thus the training data difference replicates on a second dataset of facts.

\clearpage

\section{Detecting facts after behavioral forgetting}
\label{app:access-checks}

We consider repeated statement facts that fail accuracy checks after the stream of twenty writes.
Of these, the canonical statement retains a median 69\% of
the increase in log-probability produced by its original write, as shown in Figure~\ref{fig:stored-signal-survives}b. 
This leaves
open whether the failure to answer correctly is specific to retrieving the earlier written facts
or whether it reflects broader damage to the model's ability to answer questions induced by the stream of writing. 

We make three findings to try and answer this question. 

\begin{figure}[ht]
  \centering
  \includegraphics[width=\linewidth]{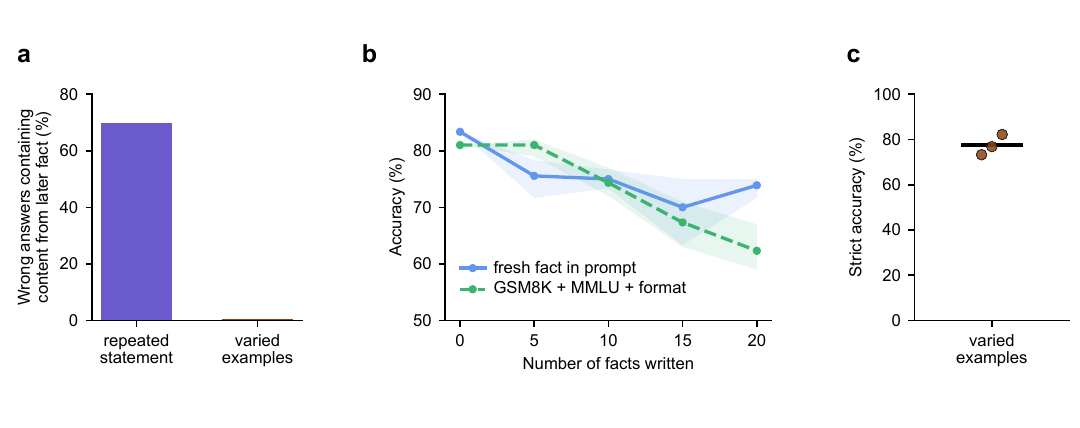}
  \caption{\textbf{Behavioral forgetting may be a selective loss of access rather
  than model damage.} \textbf{a}) Among facts failing every strict question, 68.8\% of incorrect responses after repeated statement training contained content from a later written fact (225/327 responses across 54 facts), compared with 1\% after varied-example training ($n=3$ facts). \textbf{b}) Strict accuracy on a fresh fact supplied in the prompt and accuracy on a general benchmark set across twenty sequential writes. Lines
  are means  over three seeds and shading shows the observed range across seeds.  
  \textbf{c}) Strict accuracy after twenty varied example writes on application and composition questions with the corresponding fact supplied in the prompt. Dots show accuracy per seed; the black line shows accuracy pooled across all responses and seeds (77\%).}
  \label{fig:app-access-checks}
\end{figure}

First, consider the 327 incorrect responses produced after the twentieth write for the repeated statement facts that failed every strict question. Of these responses, 225 (68.8\%) contain content from a fact written later in the sequence (Figure~\ref{fig:app-access-checks}a).
As a separate control, we supply the statement of each of 20 invented facts that never appeared in the write sequence, then ask three application or
composition questions about each fact. We also assess its performance on an unrelated 100-item benchmark comprising 40 GSM8K \citep{gsm8k}, 40 MMLU \citep{mmlu}, and 20 format-following questions. After twenty varied example writes, strict accuracy on application and composition questions decreases from 83.3\% to 73.9\%, whereas mean
benchmark accuracy decreases from 81.0\% to 62.3\%
(Figure~\ref{fig:app-access-checks}b). 
Our sequential writing therefore appears to cause broader
capability degradation that we highlight in the limitations of this work. 

Finally, after twenty varied example writes, we place each of the twenty
written fact statements back in the prompt and re-evaluate their 
application and composition questions. Pooled across seeds, strict accuracy is 77.4\%
(Figure~\ref{fig:app-access-checks}c). Panels~\textbf{b} and~\textbf{c}
therefore show that the updated model can still use facts supplied explicitly
in the prompt, despite broader performance degradation. 

\clearpage

\section{Compute resources}
\label{app:compute}

All training and evaluation jobs ran on NVIDIA GPUs. We used standalone H100 80 GB workers (one GPU each) and B200 nodes (eight GPUs each, one job per GPU). The H100 workers ran the main single-write and twenty-write experiments, the lost-access analyses (Section 3.4), and the O-LoRA comparison, each on one worker at a time; the Qwen3-8B,  full fine-tuning, and the second dataset replication (Appendix E) used four, five, and six workers in parallel. The B200 nodes ran the sixteen condition comparison (Appendix C; six nodes per phase), and the earlier-versus-later-writes experiment (Figure 3; 36 jobs, at most ten nodes at once).

\clearpage

%% file: main.bbl
\begin{thebibliography}{38}
\providecommand{\natexlab}[1]{#1}
\providecommand{\url}[1]{\texttt{#1}}
\expandafter\ifx\csname urlstyle\endcsname\relax
  \providecommand{\doi}[1]{doi: #1}\else
  \providecommand{\doi}{doi: \begingroup \urlstyle{rm}\Url}\fi

\bibitem[Agarwal et~al.(2024)Agarwal, Vieillard, Zhou, Stanczyk, Ramos~Garea, Geist, and Bachem]{agarwal2024gkd}
Rishabh Agarwal, Nino Vieillard, Yongchao Zhou, Piotr Stanczyk, Sabela Ramos~Garea, Matthieu Geist, and Olivier Bachem.
\newblock On-policy distillation of language models: Learning from self-generated mistakes.
\newblock In \emph{International Conference on Learning Representations}, pages 21246--21263, 2024.
\newblock URL \url{https://proceedings.iclr.cc/paper_files/paper/2024/file/5be69a584901a26c521c2b51e40a4c20-Paper-Conference.pdf}.

\bibitem[Allen-Zhu and Li(2024{\natexlab{a}})]{allenzhu2024physics31}
Zeyuan Allen-Zhu and Yuanzhi Li.
\newblock Physics of language models: Part 3.1, knowledge storage and extraction.
\newblock \emph{arXiv preprint arXiv:2309.14316}, 2024{\natexlab{a}}.
\newblock URL \url{https://arxiv.org/abs/2309.14316}.

\bibitem[Allen-Zhu and Li(2024{\natexlab{b}})]{allenzhu2024physics32}
Zeyuan Allen-Zhu and Yuanzhi Li.
\newblock Physics of language models: Part 3.2, knowledge manipulation.
\newblock \emph{arXiv preprint arXiv:2309.14402}, 2024{\natexlab{b}}.
\newblock URL \url{https://arxiv.org/abs/2309.14402}.

\bibitem[Askell et~al.(2021)Askell, Bai, Chen, et~al.]{askell2021lab}
Amanda Askell, Yuntao Bai, Anna Chen, et~al.
\newblock A general language assistant as a laboratory for alignment.
\newblock \emph{arXiv preprint arXiv:2112.00861}, 2021.
\newblock URL \url{https://arxiv.org/abs/2112.00861}.

\bibitem[Balesni et~al.(2024)Balesni, Korbak, and Evans]{balesni2024twohop}
Mikita Balesni, Tomek Korbak, and Owain Evans.
\newblock Lessons from studying two-hop latent reasoning.
\newblock \emph{arXiv preprint arXiv:2411.16353}, 2024.
\newblock URL \url{https://arxiv.org/abs/2411.16353}.

\bibitem[Berglund et~al.(2023)Berglund, Stickland, Balesni, Kaufmann, Tong, Korbak, Kokotajlo, and Evans]{berglund2023outofcontext}
Lukas Berglund, Asa~Cooper Stickland, Mikita Balesni, Max Kaufmann, Meg Tong, Tomasz Korbak, Daniel Kokotajlo, and Owain Evans.
\newblock Taken out of context: On measuring situational awareness in {LLMs}.
\newblock \emph{arXiv preprint arXiv:2309.00667}, 2023.
\newblock URL \url{https://arxiv.org/abs/2309.00667}.

\bibitem[Berglund et~al.(2024)Berglund, Tong, Kaufmann, Balesni, Stickland, Korbak, and Evans]{berglund2024reversal}
Lukas Berglund, Meg Tong, Max Kaufmann, Mikita Balesni, Asa~Cooper Stickland, Tomasz Korbak, and Owain Evans.
\newblock The reversal curse: {LLMs} trained on ``{A} is {B}'' fail to learn ``{B} is {A}''.
\newblock \emph{arXiv preprint arXiv:2309.12288}, 2024.
\newblock URL \url{https://arxiv.org/abs/2309.12288}.

\bibitem[Biderman et~al.(2024)Biderman, Portes, et~al.]{biderman2024lora}
Dan Biderman, Jacob Portes, et~al.
\newblock {LoRA} learns less and forgets less.
\newblock \emph{Transactions on Machine Learning Research}, 2024.
\newblock URL \url{https://arxiv.org/abs/2405.09673}.

\bibitem[Cobbe et~al.(2021)Cobbe, Kosaraju, Bavarian, et~al.]{gsm8k}
Karl Cobbe, Vineet Kosaraju, Mohammad Bavarian, et~al.
\newblock Training verifiers to solve math word problems.
\newblock \emph{arXiv preprint arXiv:2110.14168}, 2021.
\newblock URL \url{https://arxiv.org/abs/2110.14168}.

\bibitem[Gekhman et~al.(2024)Gekhman, Yona, Aharoni, et~al.]{gekhman2024finetuning}
Zorik Gekhman, Gal Yona, Roee Aharoni, et~al.
\newblock Does fine-tuning {LLMs} on new knowledge encourage hallucinations?
\newblock In \emph{EMNLP}, 2024.
\newblock URL \url{https://aclanthology.org/2024.emnlp-main.444/}.

\bibitem[Gupta et~al.(2025)Gupta, Prateepamornkul, Lu, Alaa, Hartvigsen, and Anumanchipalli]{gupta2025lifelong}
Akshat Gupta, Phudish Prateepamornkul, Maochuan Lu, Ahmed Alaa, Thomas Hartvigsen, and Gopala Anumanchipalli.
\newblock Lifelong knowledge editing requires better regularization.
\newblock \emph{arXiv preprint arXiv:2502.01636}, 2025.
\newblock URL \url{https://arxiv.org/abs/2502.01636}.

\bibitem[Hartvigsen et~al.(2023)Hartvigsen, Sankaranarayanan, Palangi, Kim, and Ghassemi]{hartvigsen2023grace}
Thomas Hartvigsen, Swami Sankaranarayanan, Hamid Palangi, Yoon Kim, and Marzyeh Ghassemi.
\newblock Aging with grace: Lifelong model editing with discrete key-value adaptors.
\newblock In \emph{NeurIPS}, 2023.
\newblock URL \url{https://openreview.net/forum?id=Oc1SIKxwdV}.

\bibitem[Hendrycks et~al.(2021)Hendrycks, Burns, Basart, Zou, Mazeika, Song, and Steinhardt]{mmlu}
Dan Hendrycks, Collin Burns, Steven Basart, Andy Zou, Mantas Mazeika, Dawn Song, and Jacob Steinhardt.
\newblock Measuring massive multitask language understanding.
\newblock In \emph{ICLR}, 2021.
\newblock URL \url{https://arxiv.org/abs/2009.03300}.

\bibitem[Holmov et~al.(2026)Holmov, Youssef, Schoots, and Seifert]{holmov2026hidden}
Ali Holmov, Paul Youssef, Nandi Schoots, and Christin Seifert.
\newblock One mask to rule them all: On hidden facts after editing and how to find them.
\newblock \emph{arXiv preprint arXiv:2605.28839}, 2026.
\newblock URL \url{https://arxiv.org/abs/2605.28839}.

\bibitem[Hu et~al.(2022)Hu, Shen, Wallis, Allen-Zhu, Li, Wang, Wang, and Chen]{hu2022lora}
Edward~J. Hu, Yelong Shen, Phillip Wallis, Zeyuan Allen-Zhu, Yuanzhi Li, Shean Wang, Lu~Wang, and Weizhu Chen.
\newblock {LoRA}: Low-rank adaptation of large language models.
\newblock In \emph{ICLR}, 2022.
\newblock URL \url{https://arxiv.org/abs/2106.09685}.

\bibitem[Jang et~al.(2022)Jang, Ye, Yang, Shin, Han, Kim, Choi, and Seo]{jang2022ckl}
Joel Jang, Seonghyeon Ye, Sohee Yang, Joongbo Shin, Janghoon Han, Gyeonghun Kim, Stanley~Jungkyu Choi, and Minjoon Seo.
\newblock Towards continual knowledge learning of language models.
\newblock In \emph{ICLR}, 2022.
\newblock URL \url{https://openreview.net/forum?id=vfsRB5MImo9}.

\bibitem[Kirkpatrick et~al.(2017)Kirkpatrick, Pascanu, Rabinowitz, et~al.]{kirkpatrick2017ewc}
James Kirkpatrick, Razvan Pascanu, Neil Rabinowitz, et~al.
\newblock Overcoming catastrophic forgetting in neural networks.
\newblock \emph{Proceedings of the National Academy of Sciences}, 114\penalty0 (13), 2017.
\newblock URL \url{https://doi.org/10.1073/pnas.1611835114}.

\bibitem[Kotha et~al.(2024)Kotha, Springer, and Raghunathan]{kotha2024implicit}
Suhas Kotha, Jacob~Mitchell Springer, and Aditi Raghunathan.
\newblock Understanding catastrophic forgetting in language models via implicit inference.
\newblock In \emph{ICLR}, 2024.
\newblock URL \url{https://openreview.net/forum?id=VrHiF2hsrm}.

\bibitem[McCloskey and Cohen(1989)]{mccloskey1989catastrophic}
Michael McCloskey and Neal~J. Cohen.
\newblock Catastrophic interference in connectionist networks: The sequential learning problem.
\newblock \emph{Psychology of Learning and Motivation}, 24:\penalty0 109--165, 1989.
\newblock \doi{10.1016/S0079-7421(08)60536-8}.
\newblock URL \url{https://doi.org/10.1016/S0079-7421(08)60536-8}.

\bibitem[Mecklenburg et~al.(2024)Mecklenburg, Lin, Li, et~al.]{mecklenburg2024injecting}
Nick Mecklenburg, Yiyou Lin, Xiaoxiao Li, et~al.
\newblock Injecting new knowledge into large language models via supervised fine-tuning.
\newblock \emph{arXiv preprint arXiv:2404.00213}, 2024.
\newblock URL \url{https://arxiv.org/abs/2404.00213}.

\bibitem[Meng et~al.(2022)Meng, Bau, Andonian, and Belinkov]{meng2022rome}
Kevin Meng, David Bau, Alex Andonian, and Yonatan Belinkov.
\newblock Locating and editing factual associations in {GPT}.
\newblock In \emph{NeurIPS}, 2022.
\newblock URL \url{https://arxiv.org/abs/2202.05262}.

\bibitem[Meng et~al.(2023)Meng, Sen~Sharma, Andonian, Belinkov, and Bau]{meng2023memit}
Kevin Meng, Arnab Sen~Sharma, Alex Andonian, Yonatan Belinkov, and David Bau.
\newblock Mass-editing memory in a transformer.
\newblock In \emph{ICLR}, 2023.
\newblock URL \url{https://arxiv.org/abs/2210.07229}.

\bibitem[Mitchell et~al.(2022)Mitchell, Lin, Bosselut, Finn, and Manning]{mitchell2022mend}
Eric Mitchell, Charles Lin, Antoine Bosselut, Chelsea Finn, and Christopher~D. Manning.
\newblock Fast model editing at scale.
\newblock In \emph{ICLR}, 2022.
\newblock URL \url{https://arxiv.org/abs/2110.11309}.

\bibitem[Padmanabhan et~al.(2023)Padmanabhan, Onoe, Zhang, Durrett, and Choi]{padmanabhan2023propagating}
Shankar Padmanabhan, Yasumasa Onoe, Michael J.~Q. Zhang, Greg Durrett, and Eunsol Choi.
\newblock Propagating knowledge updates to lms through distillation.
\newblock In \emph{NeurIPS}, 2023.
\newblock URL \url{https://proceedings.neurips.cc/paper_files/paper/2023/hash/932147114c48f8b04d41aebc0c631158-Abstract-Conference.html}.

\bibitem[Park et~al.(2025)Park, Zhang, and Tanaka]{park2025newnews}
Core~Francisco Park, Zechen Zhang, and Hidenori Tanaka.
\newblock New news: System-2 fine-tuning for robust integration of new knowledge.
\newblock \emph{arXiv preprint arXiv:2505.01812}, 2025.
\newblock URL \url{https://arxiv.org/abs/2505.01812}.

\bibitem[Patil et~al.(2024)Patil, Hase, and Bansal]{patil2024deleted}
Vaidehi Patil, Peter Hase, and Mohit Bansal.
\newblock Can sensitive information be deleted from {LLMs}? objectives for defending against extraction attacks.
\newblock In \emph{ICLR}, 2024.
\newblock URL \url{https://arxiv.org/abs/2309.17410}.

\bibitem[Shenfeld et~al.(2025)Shenfeld, Pari, and Agrawal]{shenfeld2025rlrazor}
Idan Shenfeld, Jyothish Pari, and Pulkit Agrawal.
\newblock Rl's razor: Why online reinforcement learning forgets less.
\newblock \emph{arXiv preprint arXiv:2509.04259}, 2025.
\newblock URL \url{https://arxiv.org/abs/2509.04259}.

\bibitem[Shenfeld et~al.(2026)Shenfeld, Damani, H{\"u}botter, and Agrawal]{shenfeld2026sdft}
Idan Shenfeld, Mehul Damani, Jonas H{\"u}botter, and Pulkit Agrawal.
\newblock Self-distillation enables continual learning.
\newblock \emph{arXiv preprint arXiv:2601.19897}, 2026.
\newblock URL \url{https://arxiv.org/abs/2601.19897}.

\bibitem[Snell et~al.(2022)Snell, Klein, and Zhong]{snell2022context}
Charlie Snell, Dan Klein, and Ruiqi Zhong.
\newblock Learning by distilling context.
\newblock \emph{arXiv preprint arXiv:2209.15189}, 2022.
\newblock URL \url{https://arxiv.org/abs/2209.15189}.

\bibitem[Sun et~al.(2025)Sun, Yang, Li, Li, and Gao]{sun2025pseudo}
Huashan Sun, Yizhe Yang, Yinghao Li, Jiawei Li, and Yang Gao.
\newblock Unveiling and addressing pseudo forgetting in large language models.
\newblock In \emph{Findings of ACL}, 2025.
\newblock URL \url{https://aclanthology.org/2025.findings-acl.1212/}.

\bibitem[Thede et~al.(2025)]{thede2025lifelong}
Lukas Thede et~al.
\newblock Wikibigedit: Understanding the limits of lifelong knowledge editing in {LLMs}.
\newblock In \emph{ICML}, 2025.
\newblock URL \url{https://proceedings.mlr.press/v267/thede25a.html}.

\bibitem[Wang et~al.(2024)Wang, Li, Zhang, Xu, Yao, Jiang, Xie, Huang, and Chen]{wang2024wise}
Peng Wang, Zexi Li, Ningyu Zhang, Ziwen Xu, Yunzhi Yao, Yong Jiang, Pengjun Xie, Fei Huang, and Huajun Chen.
\newblock Wise: Rethinking the knowledge memory for lifelong model editing of large language models.
\newblock In \emph{NeurIPS}, 2024.
\newblock URL \url{https://proceedings.neurips.cc/paper_files/paper/2024/hash/60960ad78868fce5c165295fbd895060-Abstract-Conference.html}.

\bibitem[Wang et~al.(2023)Wang, Chen, Ge, et~al.]{wang2023olora}
Xiao Wang, Tianze Chen, Qiming Ge, et~al.
\newblock Orthogonal subspace learning for language model continual learning.
\newblock In \emph{Findings of EMNLP}, 2023.
\newblock URL \url{https://aclanthology.org/2023.findings-emnlp.715/}.

\bibitem[Yang et~al.(2025)Yang, Li, Yang, et~al.]{qwen3}
An~Yang, Anfeng Li, Baosong Yang, et~al.
\newblock Qwen3 technical report.
\newblock \emph{arXiv preprint arXiv:2505.09388}, 2025.
\newblock URL \url{https://arxiv.org/abs/2505.09388}.

\bibitem[Yang et~al.(2024)Yang, Jones, Mozer, and Ren]{yang2024anticipatory}
Yanlai Yang, Matt Jones, Michael~C. Mozer, and Mengye Ren.
\newblock Reawakening knowledge: Anticipatory recovery from catastrophic interference via structured training.
\newblock In \emph{NeurIPS}, 2024.
\newblock URL \url{https://proceedings.neurips.cc/paper_files/paper/2024/hash/9628b97f74cd8f83ad12b631378b81e2-Abstract-Conference.html}.

\bibitem[Ye et~al.(2026)Ye, Dong, Wu, Huang, and Wei]{ye2026opcd}
Tianzhu Ye, Li~Dong, Xun Wu, Shaohan Huang, and Furu Wei.
\newblock On-policy context distillation for language models.
\newblock \emph{arXiv preprint arXiv:2602.12275}, 2026.
\newblock URL \url{https://arxiv.org/abs/2602.12275}.

\bibitem[Zhu et~al.(2020)Zhu, Rawat, Zaheer, Bhojanapalli, Li, Yu, and Kumar]{zhu2020modifying}
Chen Zhu, Ankit~Singh Rawat, Manzil Zaheer, Srinadh Bhojanapalli, Daliang Li, Felix Yu, and Sanjiv Kumar.
\newblock Modifying memories in transformer models.
\newblock \emph{arXiv preprint arXiv:2012.00363}, 2020.
\newblock URL \url{https://arxiv.org/abs/2012.00363}.

\bibitem[Zucchet et~al.(2025)Zucchet, Bornschein, Chan, Lampinen, Pascanu, and De]{zucchet2025facts}
Nicolas Zucchet, J{\"o}rg Bornschein, Stephanie Chan, Andrew~K. Lampinen, Razvan Pascanu, and Soham De.
\newblock How do language models learn facts? {Dynamics}, curricula and hallucinations.
\newblock \emph{arXiv preprint arXiv:2503.21676}, 2025.
\newblock URL \url{https://arxiv.org/abs/2503.21676}.

\end{thebibliography}
